\documentclass[
]{ceurart}

\usepackage{listings}
\usepackage{array,booktabs,multirow}
\usepackage[title]{appendix}
\usepackage{microtype}
\usepackage{graphicx}
\usepackage{adjustbox}
\usepackage{subfigure}
\usepackage{nicematrix}
\usepackage{bm}
\usepackage{svg}
\usepackage{algorithm,algpseudocode}

 \newcolumntype{M}[1]{>{\centering\arraybackslash}m{#1}}
 \newcolumntype{?}{!{\vrule width 1pt}}

\newcommand{\topf}[1]{\textbf{#1}}
\newcommand{\tops}[1]{\textbf{#1}}
\newcommand{\topt}[1]{\textbf{#1}}

\begin{document}

\copyrightyear{2022}
\copyrightclause{Copyright for this paper by its authors.
  Use permitted under Creative Commons License Attribution 4.0
  International (CC BY 4.0).}

% \conference{Woodstock'22: Symposium on the irreproducible science,
%   June 07--11, 2022, Woodstock, NY}

% \title{TTRS: Transactions Recommender System Benchmark}
\title{Next Period Recommendation Reality Check}

% \author{
%     Sergey Kolesnikov,
%     Oleg Lashinin,
%     Michail Pechatov,
%     Alexander Kosov
% }
% \affiliation{
%     Tinkoff\\
%     \{s.s.kolesnikov, o.a.lashinin, m.pechatov, a.kosov\}@tinkoff.ai\\
%     \country{Russia}
% }

% \author{Anonymous Submission}[%
% orcid=0000-0002-0877-7063,
% email=kulyabov-ds@rudn.ru,
% url=https://yamadharma.github.io/,
\author{Sergey Kolesnikov}[email=s.s.kolesnikov@tinkoff.ai]
\author{Oleg Lashinin}[email=o.a.lashinin@tinkoff.ai]
\author{Michail Pechatov}[email=m.pechatov@tinkoff.ai]
\author{Alexander Kosov}[email=a.kosov@tinkoff.ai]
% \cormark[1]
% \address[1]{Peoples' Friendship University of Russia (RUDN University),
%   6 Miklukho-Maklaya St, Moscow, 117198, Russian Federation}
% \address[2]{Joint Institute for Nuclear Research,
%   6 Joliot-Curie, Dubna, Moscow region, 141980, Russian Federation}

% \author[3]{Ilaria Tiddi}[%
% orcid=0000-0001-7116-9338,
% email=i.tiddi@vu.nl,
% url=https://kmitd.github.io/ilaria/,
% \address[3]{Vrije Universiteit Amsterdam, De Boelelaan 1105, 1081 HV Amsterdam, The Netherlands}

% \author[4]{Manfred Jeusfeld}[%
% orcid=0000-0002-9421-8566,
% email=Manfred.Jeusfeld@acm.org,
% url=http://conceptbase.sourceforge.net/mjf/,
% \address[4]{University of Skövde, Högskolevägen 1, 541 28 Skövde, Sweden}

% %% Footnotes
% \cortext[1]{Corresponding author.}
% \fntext[1]{These authors contributed equally.}

\begin{abstract}
Over the past decade, tremendous progress has been made in Recommender Systems (RecSys) for well-known tasks such as next-item and next-basket prediction. On the other hand, the recently proposed next-period recommendation (NPR) task is not covered as much. Current works about NPR are mostly based around distinct problem formulations, methods, and proprietary datasets, making solutions difficult to reproduce. In this article, we aim to fill the gap in RecSys methods evaluation on the NPR task using publicly available datasets and (1) introduce the TTRS, a large-scale financial transactions dataset suitable for RecSys methods evaluation; (2) benchmark popular RecSys approaches on several datasets for the NPR task. 
When performing our analysis, we found a strong repetitive consumption pattern in several real-world datasets.
With this setup, our results suggest that the repetitive nature of data is still hard to generalize for the evaluated RecSys methods, and novel item prediction performance is still questionable. 
\end{abstract}


\maketitle

\section{Introduction}

Recommender Systems (RecSys) have a pivotal role in improving user experience in various domains \cite{10.1145/3411174.3411194, 10.1007/978-981-13-7403-6_26, 8212834}. Furthermore, there are still new RecSys tasks being introduced, such as the recently formulated next-period recommendation (NPR) task \cite{Zhang2020HowTR, multitourrecommendations, multiperiodbasket}. This task has many real-world scenarios, including purchase planning for the near future \cite{multiperiodbasket}, 
% next-day food consumption prediction \cite{liu2019characterizing}
or travel planning \cite{multitourrecommendations}. When compared to alternatives, this task also has a different formulation. In contrast to the next-item recommendation task \cite{10.1145/3190616}, 
% NPR requires several items as a recommendation rather than only one. 
in NPR users may consume several items during the period rather than only one.
It is important to note that the next-basket recommendation task \cite{nbr22} is not appropriate here either, as users may interact with items over time separately without any formal "basket".

Despite its practical applications, the NPR task is not well-studied yet. While popular next-item and next-basket tasks are actively benchmarked on publicly available datasets \cite{nbr22, 10.1145/3383313.3418479, 44265d57d70f4725bf739c67800a5f77}, NPR studies are mostly focused on specific tasks, methods and publicly unavailable datasets \cite{Zhang2020HowTR, multitourrecommendations, multiperiodbasket}. On the other hand, the NPR task could be reframed from well-known next-item and next-basket ones. Because of this, it can be possible to adapt and compare the previously proposed solutions for next-item and next-basket tasks in the same manner. Coupled with recent advances in next-item and next-basket offline evaluation \cite{10.1145/3383313.3418479, 44265d57d70f4725bf739c67800a5f77, DBLP:journals/corr/abs-2010-11060}, we see a gap in popular RecSys methods performance evaluation on the NPR task with publicly available datasets.
Furthermore, while there are various datasets for RecSys which give an overview of e-commerce or movies domains, such as MovieLens\cite{movielens}, Instacart,
% \footnote{https://www.kaggle.com/c/instacart-market-basket-analysis}
Yelp,
% \footnote{https://www.yelp.com/dataset}
TaFeng,
% \footnote{https://www.kaggle.com/datasets/chiranjivdas09/ta-feng-grocery-dataset}
and Dunnhumby,
% \footnote{https://www.dunnhumby.com/careers/engineering/sourcefiles}
there is a lack of reproducible\footnote{Following \cite{arewereallymakingmuchprogress}, we did not consider a paper reproducible if at least one dataset used in the paper was found unavailable.} research studies in other equally important domains, including personalized financial recommendations\cite{bendreGPRGlobalPersonalized2021}.

To this end, in this paper, we not only benchmark popular RecSys methods on the NPR task but also introduce a large-scale financial transactions dataset for evaluating RecSys approaches. 
To be more concrete, the contributions of this paper can be  summarized as follows:
\begin{itemize}
\item Our first contribution is a large-scale financial transactions dataset – \mbox{TTRS (\hyperref[sec:datasets]{Section 3})}. It contains over 2 million interactions between almost 10,000 users and more than 1,000 merchants for a total of 14 months. To the best of our knowledge, it is the first RecSys dataset that represents users' full financial activity rather than one with a single merchant subset.
\item Our second contribution is a benchmark of various RecSys methods on the next-period recommendation task (\hyperref[sec:experiments]{Section 4}).
To the best of our knowledge, no prior studies have evaluated these RecSys methods together on the same task.
\item Our last contribution is a purchase pattern analysis of the reviewed datasets as well as a RecSys methods' forecast performance analysis on these datasets. We found that generalizing to a strong repetitive consumption trend of real-world datasets is still problematic for current RecSys approaches and novel items prediction is still difficult to achieve under such a setup.
% \textit{propose several directions to overcome these limitations} (\hyperref[sec:discussion]{Section 5}).
\end{itemize}

% \section{Related Work}
% \section{Preliminaries}
\section{Background}

\subsection{Next-period Recommendations}
\label{sec:tasks}

There are several specific ways we can formulate recommendation tasks.
For example, suppose we are interested in forming general recommendations, such as proposing a new movie based on those already watched. In this case, we could formulate this as a top-n recommendation task \cite{conventionalrecs} and use methods that are appropriate for it: NMF \cite{nmf}, EASE \cite{ease}, or MultiVAE \cite{multivae}. On the other hand, if we are interested in predicting the next item the user will interact with, such as the next song in a playlist, it would be a next-item prediction task \cite{10.1145/3190616}. In this case, we could use next-item methods: GRU4Rec \cite{gru4rec}, Bert4Rec \cite{bert4rec}, etc. Alternatively, if a user can consume entire sets of items simultaneously and we want to predict a whole set of their interactions, such as the content of their next shopping cart in a shop, this can be formulated as a next-basket prediction task \cite{nbr22}. For this task, methods such as TifuKNN \cite{tifuknn} are usually used. Finally, if we want to determine user interests over time, we could try to predict all user interactions for a predefined period, which would make this a next-period recommendation task \cite{Zhang2020HowTR, multitourrecommendations, multiperiodbasket}. This type of task is what we are focused on in this work.
Thanks to the time-aware nature of the task, we can form interaction matrices from past periods and predict future interaction using general ton-n recommenders. Alternatively, we can order all user interactions as a sequence and use next-item approaches. Finally, we can group user interactions into period-based baskets and adapt next-basket methods. Due to such adaptive reframing, we can use methods from all discussed formulations and compare them on the same task.

\clearpage

\subsection{Datasets}
\label{sec:datasets}

Following recent studies \cite{10.1145/3383313.3418479}, we reviewed and compared several well-known time-aware alternatives with the proposed TTRS dataset:
% Following recent studies \cite{10.1145/3383313.3418479}, we review and compare several well-known time-aware alternatives with the proposed TTRS dataset:

\begin{itemize}
    \item Ta-Feng\footnote{\url{https://www.kaggle.com/chiranjivdas09/ta-feng-grocery-dataset}} - a Chinese grocery store dataset that has basket-based transaction data from November 2000 to February 2001.
    This dataset is widely used for next-basket prediction research \cite{tifuknn, 10.1145/3292500.3330979, ijcai2019-389}.
    \item Dunnhumby\footnote{\url{https://www.dunnhumby.com/source-files/}} (DHB) - a dataset provided by Dunnhumby that contains customers' transaction data over a period of 117 weeks from April 2006 to July 2008. For benchmarking purposes, we select the "Let's Get Sort-of-Real sample 50K customers" version of the dataset, which is well-known within the research community \cite{tifuknn, 10.1145/3292500.3330979, DBLP:journals/corr/abs-1905-11691}.
\end{itemize}
Statistical information on the raw datasets is summarized in \autoref{tab:datastat}.

\begin{table*}[!tbp]%[ht]
\scriptsize
  \centering
    \begin{tabular}{c?c|c|c?c|c|c?c|c|c|c} \toprule
    &\multicolumn{3}{c?}{Before preprocessing}&\multicolumn{3}{c?}{After preprocessing}&\multicolumn{4}{c}{Final statistics}\\ \cmidrule{2-11}
    \multirow{2}{*}{Dataset}&\multirow{3}{*}{\# users}&\multirow{3}{*}{\# items}&\multirow{3}{*}{\# inter.}&\multirow{3}{*}{\# users}&\multirow{3}{*}{\# items}&\multirow{3}{*}{\# inter.}&\# inter.&\# inter. &\# inter. & \# inter. \\ 
      &&&&&&& per user        & per user       & per item       & per item\\         
          &&&&&&&&  per period &        & per period\\  \midrule
    % TaFeng    & 32266    & 23812    & 817741&3470    & 2929    & 196549& 56 & 14 & 67 & 16     \\
    % TTRS      & 50000    & 2873     & 14287287& 9396    & 1157    & 2744828& 292 & 20 & 2372 & 169    \\ 
    % DHB & 50000    & 4997     & 31057875&11047   & 3178    & 11594609& 1049 & 40 & 3648 & 140   \\
    TaFeng    & 32266    & 23812    & 0.8m & 3470    & 2929    & 0.2m & 56 & 14 & 67 & 16     \\
    TTRS      & 50000    & 2873     & 14m & 9396    & 1157    & 3m & 292 & 20 & 2372 & 169    \\ 
    DHB & 50000    & 4997     & 31m & 11047   & 3178    & 12m& 1049 & 40 & 3648 & 140   \\
   	\bottomrule
    \end{tabular}
  \caption{Dataset statistics before and after preprocessing. We measured the number of users, items, and interactions. For overall statistics, we counted interactions in millions (m). For final statistics, we used one month as a period.}

\label{tab:datastat}
\centering

\end{table*}

\subsection{Evaluation}

Evaluating RecSys can be challenging due to various possible data splitting strategies and data preparation approaches. 
In early works \cite{10.1145/2043932.2043990, timetoconsidertime}, the researchers highlighted the importance of time-aware algorithm validation. 
In a prior study \cite{daisyrec}, the authors sampled 85 papers published in 2017-2019 from top conferences and concluded that random-split-by-ratio and leave-one-out splitting strategies are used in 82\% of cases. 
At the same time, recent studies \cite{10.1145/3383313.3418479} pointed out that the most strict and realistic setting for data splitting is a global temporal split, where a fixed time point separates interactions for training and testing. The authors found that only 2 of 17 recommender algorithms (published from 2009 to 2020) were evaluated using this scenario. 
In another work \cite{DBLP:journals/corr/abs-2010-11060}, the authors compare the impact of data leaks on different RecSys methods. They found that "future data" can improve or deteriorate recommendation accuracy, making the impact of data leakage unpredictable. 
To avoid the issues listed above, we use a global temporal K-fold validation scheme in this paper (Section 4), applying ideas from \cite{10.1145/3383313.3418479, 44265d57d70f4725bf739c67800a5f77, DBLP:journals/corr/abs-2010-11060}.

\section{Dataset}

\begin{figure}[t!]
    \centering
    \begin{subfigure}[]
        \centering
        \includegraphics[width=0.49\textwidth]{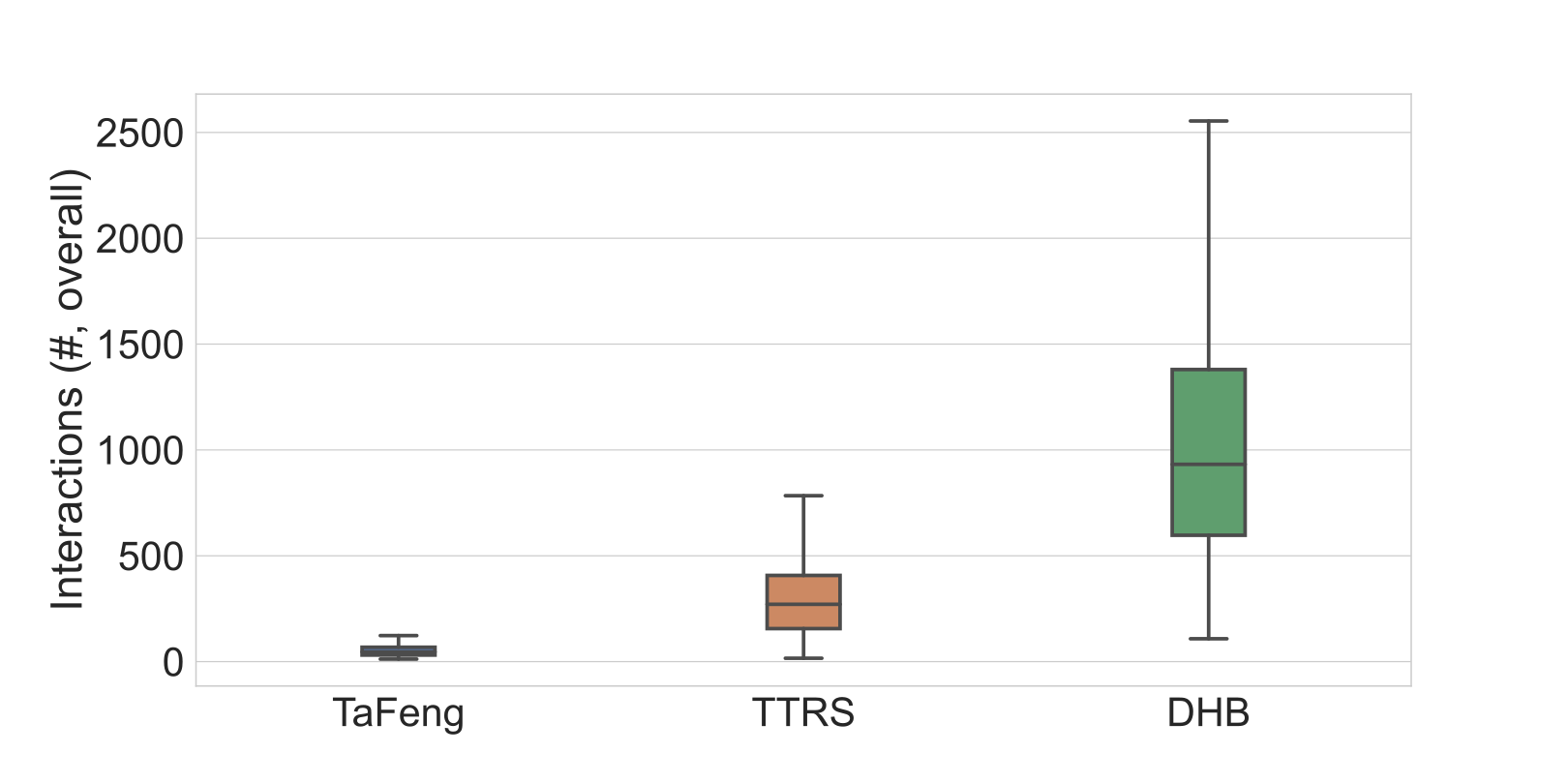}
    \end{subfigure}
    \begin{subfigure}[]
        \centering
        \includegraphics[width=0.49\textwidth]{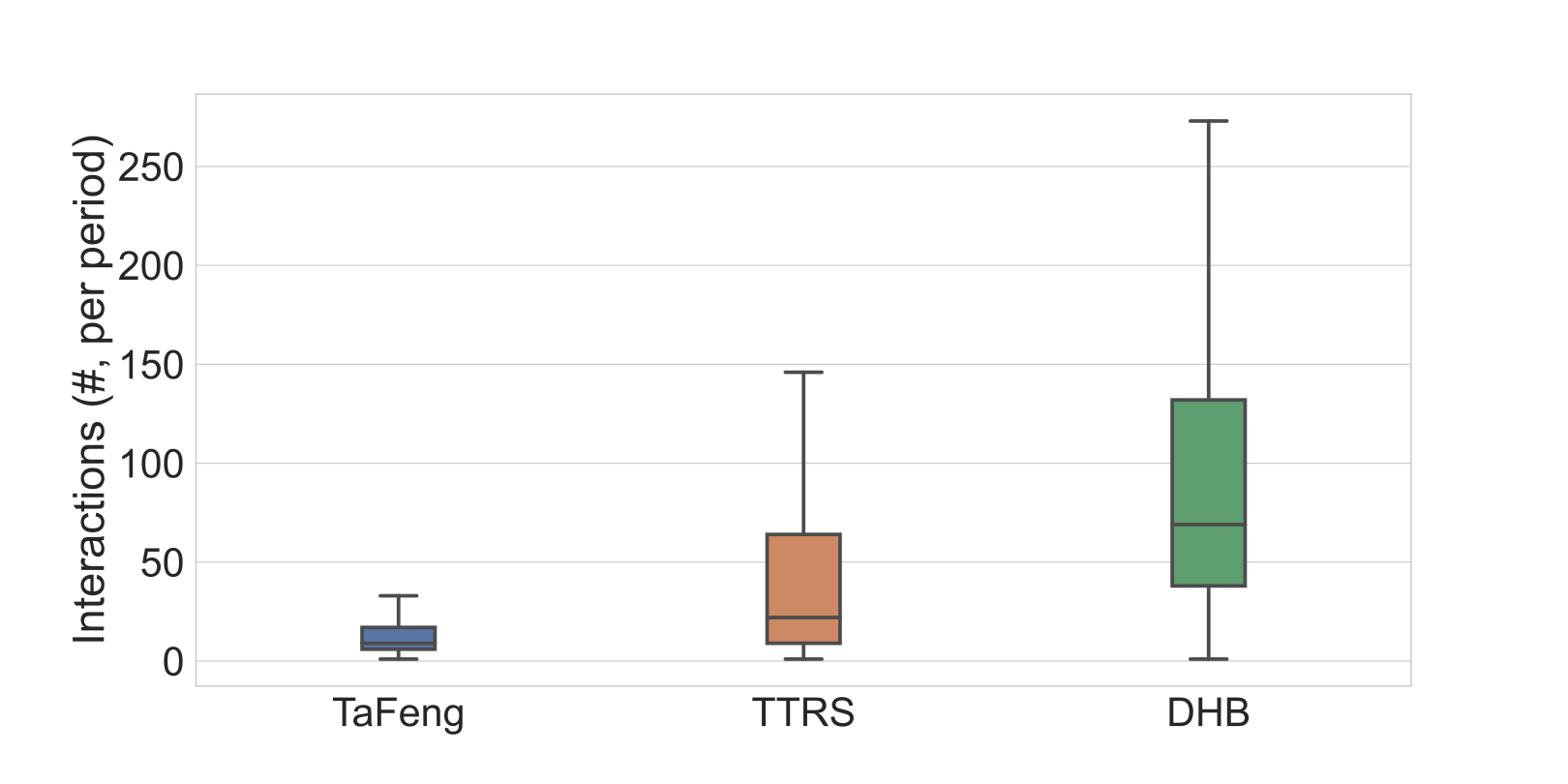}
    \end{subfigure}
    \begin{subfigure}[]
        \centering
        \includegraphics[width=0.49\textwidth]{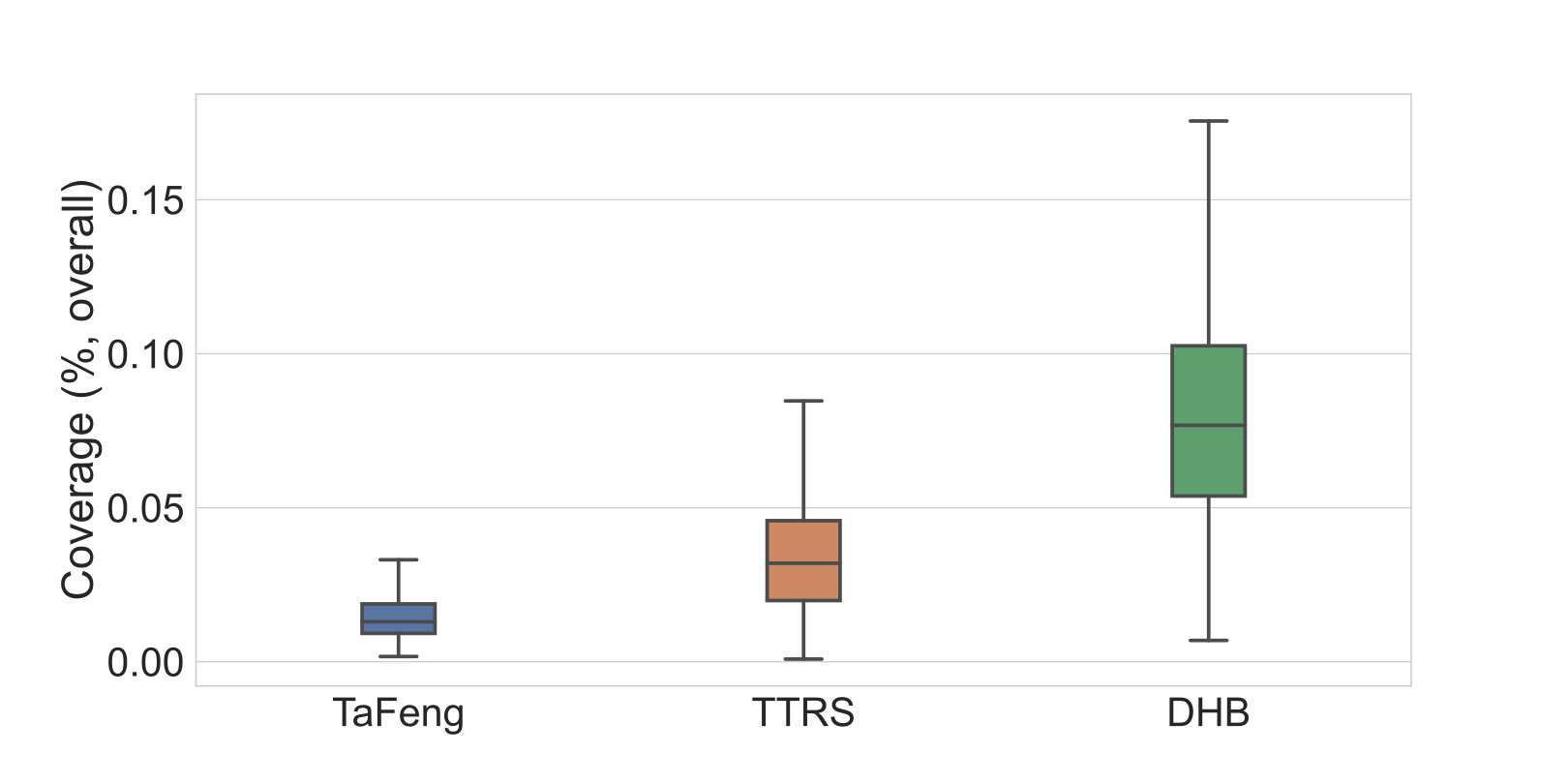}
    \end{subfigure}
    \begin{subfigure}[]
        \centering
        \includegraphics[width=0.49\textwidth]{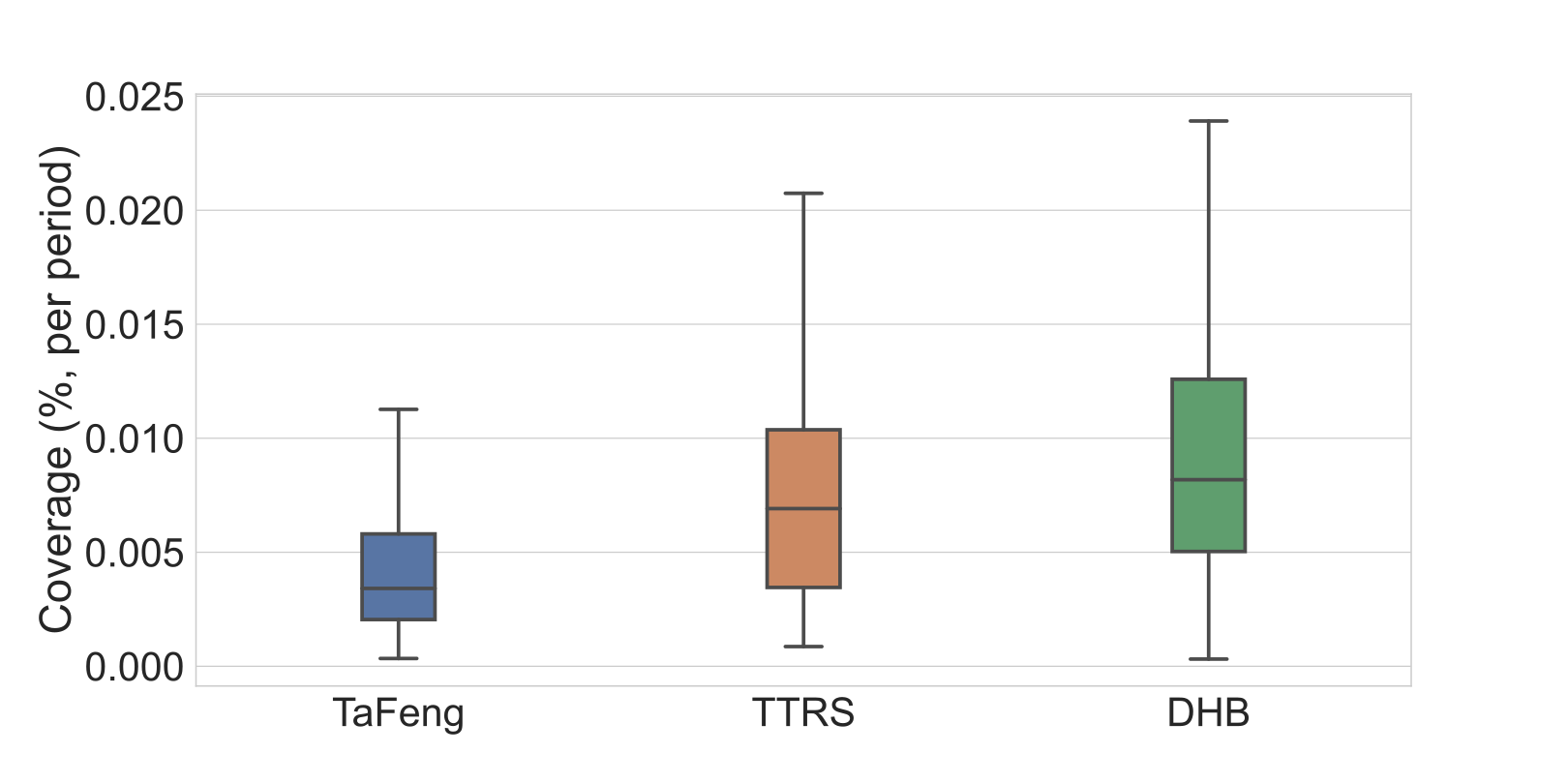}
    \end{subfigure}
    \caption{Dataset statistics: (a) overall number of user interactions, (b) number of user interactions per period, (c) items' overall coverage, (d) items' coverage per-period. We used one month as a period for each dataset.}
    \label{fig:data-comparison}
\end{figure}

Our TTRS dataset\footnote{The dataset will be available under the CC-BY-NC-SA-4.0 license after publication.} contains transaction information of a randomly selected 50 thousand \textit{T-Bank}\footnote{Anonymized.} users from January 2019 to March 2020. 
Each transaction contains an anonymized user id, transaction type, merchant information, transaction timestamp and amount. Full description of the dataset can be found under \hyperref[tab:DATASETDESCRIPTION]{Appendix}. 

% The key property of the TTRS dataset is how diverse its transaction sources are.
The key property of the TTRS dataset is the diversity of the presented transaction sources.
While other datasets mainly provide users' activity with a single merchant, TTRS contains users' entire financial activity, including interactions with different supermarkets, clothing stores, online delivery services, cinemas, gas stations, cafes and restaurants, museums, etc. Therefore, TTRS contains anonymized information about users' daily interests based on their transactional activity. 
As a result, in contrast to other available datasets, TTRS consists of both online and offline purchases.

\textbf{Preprocessing Pipeline.} To reduce possible anomalies, we 
% remove users and items with less than ten interactions in the first six months and 
filter out users with less than one transaction per month, similar to \cite{fpmc}. As filtering can change the number of interactions between the remaining users and items, we repeat this step several times until our data converges. Statistical information about datasets after preprocessing is summarized in \autoref{tab:datastat}.

\begin{figure}
    \centering
    \begin{subfigure}[]
        \centering
        \includegraphics[width=0.49\textwidth]{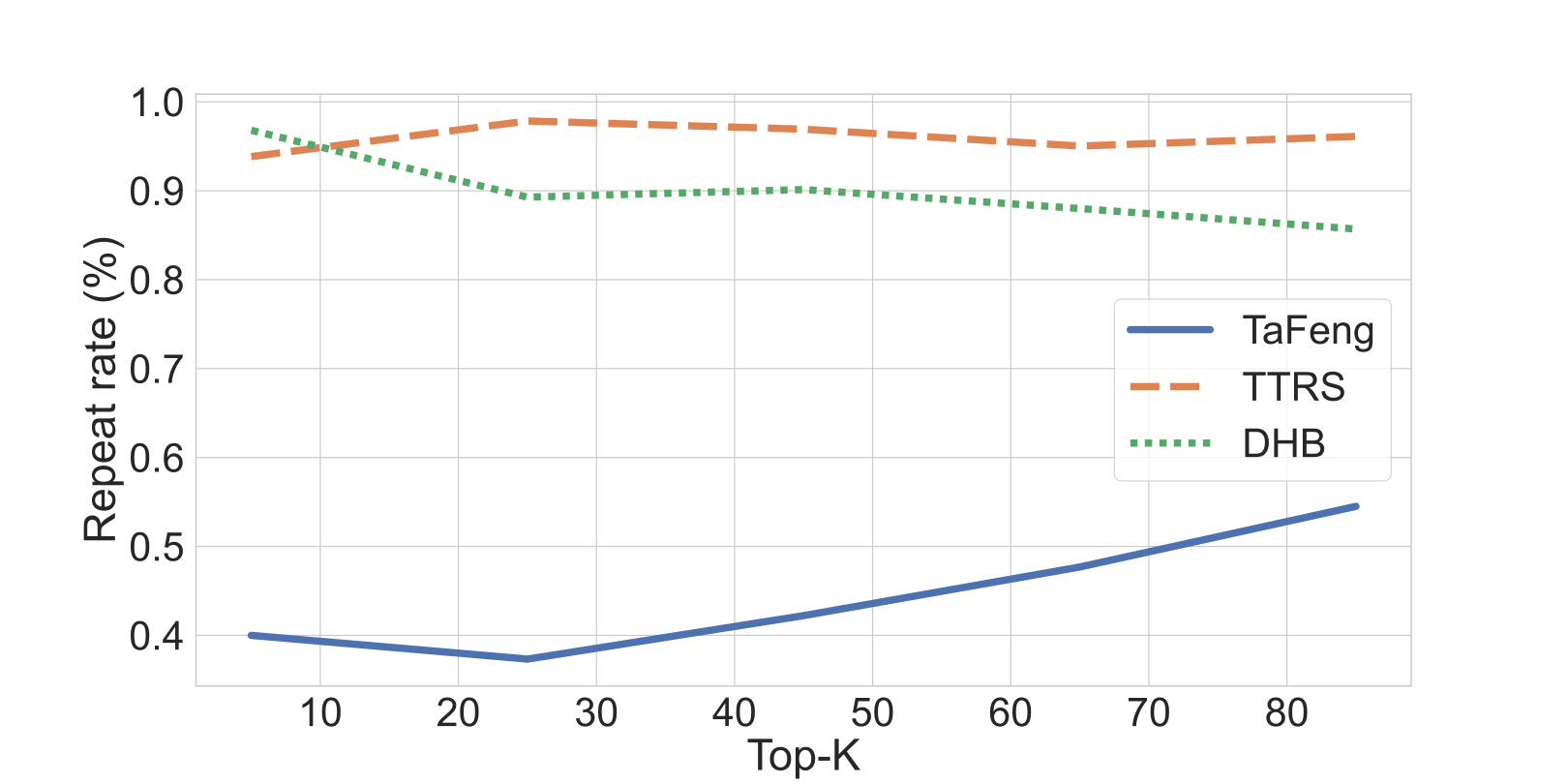}
        % \label{fig:data-repeat-score}
    \end{subfigure}
    \begin{subfigure}[]
        \centering
        \includegraphics[width=0.49\textwidth]{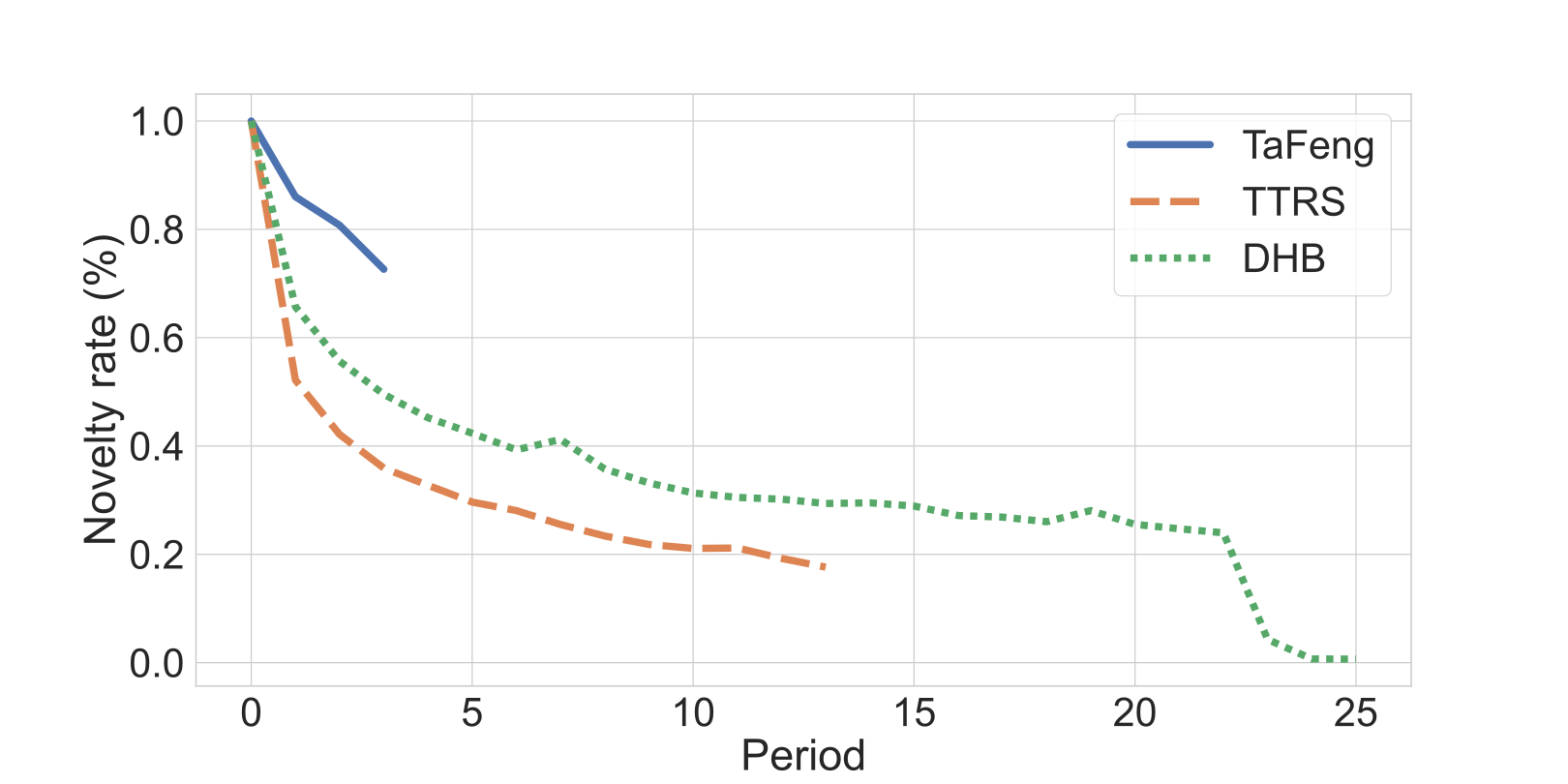}
        % \label{fig:data-new-score}
    \end{subfigure}
    \caption{
    \newline(a) Repeat rate of popular items. For each Top-K (X-axis), we measure the mean top-k-popular items purchase repeat score (Y-axis) over all dataset periods. We use one month as a period for each dataset. 
    \newline(b) Novelty rate of items. For each period (X-axis), we measure the mean novelty of items purchased by users (Y-axis). We use one month as a period for each dataset.  
    % There are 4, 14, and 26 months in TaFeng, TTRS, and DHB datasets, respectively.
    }
    \label{fig:data-scores}
\end{figure}

\subsection{Comparison}

% \subsubsection*{User Statistics}
% \textbf{User Statistics.} 
When comparing RecSys datasets, one possible direction of inquiry can concern the difference in user item purchasing and coverage. To be more exact, we are interested in knowing the following: (1) How many items has the user purchased in total? (2) From all available items, how many unique ones has the user interacted with? As we focus on the next-period recommendation task, the same questions could be asked in terms of statistics per period.

The results of dataset comparison are presented in \autoref{fig:data-comparison}.
% , and based on these results, several observations can be made. 
% First, the 
TaFeng represents the most "sparse" dataset, as it has the lowest number of interactions and item coverage. On the other hand, DHB is the most "dense" dataset, with its statistics having the most significant values among competitors. 
We suspect that the main reason for such explicit fragmentation is the overall size of the dataset in terms of time, items, users, and diversity of available goods. While DHB contains transactions from $760$ stores over 2 years, TaFeng represents only one over 4 months.
Finally, while the proposed TTRS dataset has average overall statistics, its per period coverage is comparable with DHB, indicating a high diversity of user interests over time.

% \subsubsection*{Popular Items Analysis}
\textbf{Popular Items Analysis.}
To better understand the differences between datasets in terms of per period statistics, we additionally review how often the most popular items purchased in one period are popular in the next one.
The analysis results are presented in \autoref{fig:data-scores} (a). 
While TaFeng results gradually increase and DHB decrease with the $Top-K$ threshold, TTRS values never drop under $95\%$. 
According to these results, we can see a strong repetitive consumption pattern in TTRS and DHB datasets.

% \newpage
% \\
% \newline{}
% \hfill \break

% \subsubsection*{Novel Items Analysis}
\textbf{Novel Items Analysis.}
% Another question that we are interested in 
Alternative question that we could be interested in researching is how often users purchase new items that have not been purchased by them before. The results answering this question are presented in \autoref{fig:data-scores} (b).
According to our results, we can see that an item's novelty rate exponentially decreases with time across all datasets. 
% As the number of purely novel items decreases with every period, these results also highlight the importance of repetitive pattern analysis for modeling long-term user preferences. 
The only anomaly seen in the results is a second drop for the DHB dataset. While we find this interesting in terms of analyzing the DHB dataset, its further investigation can be seen as out of the scope of this paper.

\clearpage

\begin{figure*}[t]
    \centering
    \includegraphics[width=1.0\textwidth]{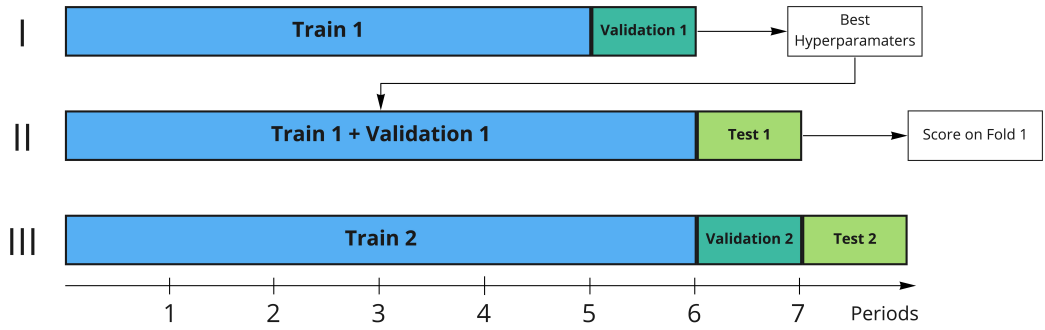}
    \caption{Evaluation procedure. On step (I), we train model on the $("Train\:1", "Validation\:1")$ split to find optimal hyperparameters for the first fold. On step (II), we use these hyperparameters to train the model on the $("Train\:1 + Validation\:1", "Test\:1")$ split to evaluate the model on the first fold. On step (III), we repeat the (I) and (II) steps for the next fold.}
    \label{fig:data_split}
\end{figure*}  

% \section{Experiments}
\section{Benchmark}
\label{sec:experiments}

For our benchmark, we chose the next-period recommendation task with a period of one month. The main goal of our benchmark was to predict all user interactions in the next month by using their interaction history over the past few months.

\subsection{Evaluation Setup}
\label{sec:evaluation-setup}

% \subsubsection*{Task}
% -

% \subsubsection*{Metrics}
\textbf{Metrics.}
We compare models with each other using standard metrics, such as $Recall@K$, $NDCG@K$, and $MAP@K$. Each metric can be calculated for a recommendation list of length K, where K ranges from 1 to the number of items. K is usually called the cutoff, which stands for the length to which the recommendations are cut. 
We use $@10$ cutoff during benchmarking
% because there are limits for showing recommendations to users
\footnote{$@20$, $@50$ and $@100$ cutoff results could be found under Appendix.}.

% \subsubsection*{Validation Scheme.}
% \label{sec:validscheme}
\textbf{Validation Scheme.}
For validation, we use several approaches from prior articles \cite{arewereallymakingmuchprogress, 10.1145/3383313.3418479,  timetoconsidertime}. Firstly, we apply a global temporal split to separate our training and test data. Secondly, we use temporal cross-validation with several folds. Finally, for each such fold with $N$ periods, we use the optimal hyperparameter search through extra data partitioning into $"train"$ (N-2 periods), $"validation"$ (1 period), and $"test"$ (1 period) splits. The best hyperparameters found on the $("train", "validation")$ split were used to initialize and train the model for final testing on the $("train+validation", "test")$ split. The entire validation process is shown in \autoref{fig:data_split}.

% \subsubsection*{Hyperparameter Search}
\textbf{Hyperparameter Search.}
Similar to previous studies \cite{arewereallymakingmuchprogress}, we search for the optimal parameters via Bayesian search using Optuna\footnote{https://optuna.org}. For each pair (algorithm and test fold), we iterate over 25 hypotheses\footnote{Hyperparameters grid can be found under Appendix.}. We use $MAP@10$ metric for model selection.

\subsection{Methods}

During our benchmark, we evaluated the following RecSys approaches: 
\begin{itemize}
    \item \textbf{TopPopular} \cite{popularbaselines}, \textbf{TopPersonal}  \cite{tifuknn} are simple RecSys popularity-based baselines.
    % that work based on item popularity. 
    % The TopPopular algorithm recommends the most popular items, sorting them in descending order of global popularity. The TopPersonal method focuses on items with which the user has already interacted. The recommendation list is created by sorting them in descending order of interaction frequency. If no personal recommendations are found, TopPersonal uses the TopPopular approach for prediction.
    \item \textbf{NMF} \cite{nmf}, \textbf{PureSVD} \cite{conventionalrecs}, \textbf{ALS} \cite{als} are matrix factorization-based (MF-based) models. 
    These models are designed to approximate any value in the interaction matrix by multiplying the user and item vectors in the hidden space. The interaction matrix could be represented as a matrix with interaction frequencies or in binary form. We use the $preprocessing$ hyperparameter to handle this data transformation.
    We apply the $log (1 + p)$ transformation for the $log$ option. For $binary$, all frequencies above 0 were replaced by 1. Finally, no transformations were used at all for the $none$ preprocessing.
    \item \textbf{SLIM} \cite{slim}, \textbf{EASE} \cite{ease} are linear models that learn an item-item weight matrix (I2I-based). Similar to MF-based models, we add a $preprocessing$ hyperparameter to approximate frequencies or a binary mask of interactions.
    \item \textbf{MultiVAE} \cite{multivae}, \textbf{MacridVAE} \cite{macridvae}, \textbf{RecVAE} \cite{recvae} are variational autoencoder approaches (VAE-based) for a top-n recommendation task. As a more generic benchmark, we also include the $preprocessing$ hyperparameter for the VAE-based methods.
    % They utilize the idea of using multinomial likelihood to recover inputs from hidden representations and use them for recommendations.
    \item \textbf{GRU4Rec} \cite{gru4rec}, \textbf{SASRec} \cite{sasrec}, \textbf{BERT4Rec} \cite{bert4rec} are sequential-based models. Unlike previously mentioned methods, these models know the sequence of users' interactions and learn sequence representation with RNN- or self-attention-based neural networks. This representation is then used for next-item recommendations.
    \item \textbf{RepeatNet} \cite{repeatnet} is an RNN-based model that uses a repeat-explore mechanism for %session-based
    recommendations. 
    This model has two different recommendation modes. In the first, "repeat" mode, the model recommends something from users' consumption history. In the second, "explore" mode, the model recommends something new that has not been listed in the input sequence. 
    \item \textbf{ItemKNN} \cite{DBLP:conf/www/SarwarKKR01} is an item-based k-nearest neighbors method, which utilizes similarities between previously purchased items. 
    Similar to \cite{arewereallymakingmuchprogress}, we used different similarity measures for our experiments: Jaccard, Cosine, Asymmetric Cosine, and Tversky similarity.
    \item \textbf{TIFUKNN} \cite{tifuknn} is a next-basket recommender which applies the idea of learning Temporal Item Frequencies with the k-nearest neighbors approach.
    % is a current state-of-the-art method for the next-basket recommendation task. It applies the idea of learning Temporal Item Frequencies with the k-nearest neighbors approach.
\end{itemize}

To summarize, we consider 16 models, 7 of which are neural networks, and 5 are sequence-aware. For benchmarking purposes, we include approaches of different types such as matrix factorization, linear models, variational autoencoders, recurrent neural networks, and self-attention-based methods.
% We adapted all the above models for the next-period recommendation task. 
Each model produced predictions in the form of sorted lists of items for the next period of one month. The models were all evaluated on the next-period recommendation task in the same manner.
% To predict the next period,
To train and evaluate all of the above models, we follow the preprocessing procedure of previous works \cite{bert4rec, sasrec} and use only 30, 40, and 80 last interactions for TaFeng, TTRS, and DHB datasets, respectively, i.e., roughly proportional to the mean number of interactions per user in a single period (\autoref{tab:datastat}).

\clearpage

\begin{table*}[t!]
\caption{Next-month recommendation benchmark. All metrics are averaged over 2 (TaFeng) or 6 (TTRS and DHB) test folds and normalized to the $[0, 100]$ values range for presentation convenience.
The top three models are marked in \textbf{bold}.
% The best model is marked in \topf{bold}, the second best model is \tops{underlined} and the third best model is in \topt{italics}.
}
\label{tab:benchmark-ml}
\begin{center}
% \begin{small}
\begin{sc}
% \footnotesize
\begin{tabular}{l|ccc|ccc|ccc}
\toprule

\multirow{2}{*}{Method} & \multicolumn{3}{c|}{TaFeng} & \multicolumn{3}{c|}{TTRS} & \multicolumn{3}{c}{DHB} \\
& Recall & MAP & NDCG & Recall & MAP & NDCG & Recall & MAP & NDCG \\
\midrule

toppopular & 5.26  & 3.34  & 7.32  & 31.71  & 20.70  & 34.69  & 9.45  & 16.20  & 29.38 \\
toppersonal & 12.51 & \topt{8.99} & \topt{16.97} & 58.68 & 52.14 & 66.28 & \tops{21.82} & \tops{42.82} & \tops{57.52}\\
\midrule
nmf & 11.33 & 7.80 & 15.31 & 57.66 & 51.43 & 65.59 & 19.13 & 38.39 & 53.40\\
svd & 12.60 & 8.54 & 16.54 & 57.81 & 51.65 & 65.77 & 19.87 & 39.63 & 54.61\\
als & 12.46 & 7.11 & 14.73 & 57.04 & 42.62  & 57.84  & 17.07 & 28.84 & 44.10\\
\midrule
ease & \tops{13.23} & \tops{9.07} & \tops{17.14} & 40.32  & 27.89  & 43.11  & 14.63  & 23.57  & 39.04 \\
slim & 11.62 & 8.22 & 15.56 & 37.71  & 25.38  & 40.30  & 12.89  & 20.33  & 35.19 \\
\midrule
multivae & 10.53  & 5.90  & 12.50  & 55.70  & 45.15  & 59.98  & 15.80 & 26.52 & 41.87\\
macridvae & 7.83  & 4.70  & 10.04  & 39.27  & 28.89  & 43.73  & 9.88  & 16.41  & 29.77 \\
recvae & 12.60 & 8.91 & 16.73 & \tops{59.65} & \tops{52.89} & \tops{66.93} & \topt{21.14} & \topt{40.52} & \topt{55.43}\\
\midrule
gru4rec & 4.18  & 2.20  & 5.19  & 55.72  & 49.66 & 63.96 & 14.61  & 24.67  & 40.19 \\
sasrec & 7.71  & 5.07  & 10.09  & 57.35 & 50.61 & 64.81 & 15.62  & 26.03  & 41.56 \\
bert4rec & 7.36  & 4.75  & 9.69  & 56.54 & 49.72 & 64.03 & 16.23 & 28.56 & 44.34\\
% \midrule
repeatnet & \topf{13.44} & \topf{9.08} & \topf{17.45} & \topf{59.76} & \topf{53.47} & \topf{67.48} & \topf{22.26} & \topf{44.08} & \topf{58.74}\\
\midrule
itemknn & 9.14  & 5.59  & 11.34  & 28.43  & 19.02  & 32.08  & 10.44  & 16.66  & 30.24 \\
tifuknn & \topt{12.88} & 8.80 & 16.80 & \topt{59.03} & \topt{52.19} & \topt{66.38} & 20.12 & 37.98 & 53.27\\

\bottomrule
\end{tabular}
\end{sc}
% \end{small}
\end{center}
\end{table*}

\subsection{Results}

The results of the models' comparison can be found in \autoref{tab:benchmark-ml}. 
TopPopular, which is the most straightforward and commonly adopted RecSys baseline, achieved minor results across all datasets. Despite such results, its personalized improvement, TopPersonal, achieved performance comparable with other methods and mostly falls into the top 4 best methods over all datasets. Many current RecSys methods of different types (MF, I2I, VAE, sequential) could hardly beat the TopPersonal baseline.

The only deep learning approach that consistently outperforms any other method is the recently proposed RepeatNet. Another notable deep learning approach is RecVAE, and while it could hardly beat the TopPersonal baseline on every dataset, it generally achieves comparable performance. 
Finally, when it comes to methods that do not utilize deep learning, TIFUKNN results are close to the TopPersonal ones, or even better in the case of the TTRS dataset.

\clearpage

\begin{table*}[t!]
\caption{Next-month novel items recommendation benchmark. Ground truth objects are new to users, and recommendation lists do not contain items interacted with during training. All metrics are averaged over 2 (TaFeng) or 6 (TTRS and DHB) test folds and normalized to the $[0, 100]$ values range for presentation convenience. 
The top three models are marked in \textbf{bold}.
% The best model is marked in \topf{bold}, the second best model is \tops{underlined} and the third best model is in \topt{italics}.
}
\label{tab:benchmark-ml-new}
\begin{center}
% \begin{small}
\begin{sc}
% \footnotesize
\begin{tabular}{l|ccc|ccc|ccc}
\toprule

\multirow{2}{*}{Method} & \multicolumn{3}{c|}{TaFeng} & \multicolumn{3}{c|}{TTRS} & \multicolumn{3}{c}{DHB} \\
& Recall & MAP & NDCG & Recall & MAP & NDCG & Recall & MAP & NDCG \\
\midrule

toppopular & 4.23 & \topt{2.80} & \topt{5.82} & 20.09  & 9.04  & 16.06  & 4.82 & 3.92 & 9.31\\
toppersonal & 4.25 & \tops{2.81} & \tops{5.85} & 20.14  & 9.11  & 16.15  & 4.77 & 3.88 & 9.22\\
\midrule
nmf & 1.36  & 0.71  & 1.73  & 3.35  & 1.34  & 2.59  & 1.65  & 1.04  & 2.87 \\
svd & 1.48  & 0.69  & 1.72  & 2.37  & 0.93  & 1.87  & 1.57  & 1.01  & 2.77 \\
als & 2.34  & 1.04  & 2.56  & 8.80  & 3.68  & 6.85  & 3.29  & 2.20  & 5.71 \\
\midrule
ease & 3.81 & 2.03 & 4.58 & \topt{23.51} & \topt{11.19} & \topt{19.02} & 4.42  & 3.36 & 8.20\\
slim & 3.62 & 1.73 & 4.14 & 22.79 & 10.62 & 18.30 & \tops{5.16} & 3.92 & \topt{9.41}\\
\midrule
multivae & 2.70  & 1.16  & 2.93  & \topf{24.02} & \tops{11.20} & \topf{19.22} & 4.74 & 3.56 & 8.68\\
macridvae & \topf{4.77} & 2.18 & 5.14 & 22.61 & 10.25 & 17.99 & \topt{5.04} & \tops{4.02} & \tops{9.58}\\
recvae & 3.77 & 1.84 & 4.34 & 23.79 & \topf{11.21} & \tops{19.18} & 4.67 & 3.65 & 8.79\\
\midrule
gru4rec & 2.60  & 1.05  & 2.79  & 21.24 & 9.25  & 16.41  & 3.85  & 2.71  & 6.86 \\
sasrec & 2.95 & 1.28  & 3.14  & 23.47 & 10.70 & 18.48 & 3.91  & 2.69  & 6.83 \\
bert4rec & 2.92  & 1.26  & 3.15 & 22.31 & 10.00 & 17.43 & 4.52 & 3.31  & 8.18 \\
% \midrule
repeatnet & \topt{4.36} & 1.37 & 3.84 & 20.65  & 9.32  & 16.50  & 2.23  & 1.67  & 4.40 \\
\midrule
itemknn & 2.71  & 1.30 & 3.08  & 22.92 & 10.72 & 18.55 & \topf{5.34} & \topf{4.25} & \topf{10.00}\\
tifuknn & \tops{4.56} & \topf{2.86} & \topf{6.03} & 21.22  & 9.67 & 16.97 & 4.84 & \topt{3.93} & 9.32\\

\bottomrule
\end{tabular}
\end{sc}
% \end{small}
\end{center}
\end{table*}

\subsubsection*{Novel Item Performance}
As an alternative direction for evaluation, we also compare the recommenders' performance in predicting completely novel items for the user.
Due to item novelty rate in user interactions decreasing with time (\autoref{fig:data-scores} (b)), we wonder how well current RecSys methods could predict novel options for the user. The results of model comparison on this task can be found in \autoref{tab:benchmark-ml-new}. 

In our comparison, we can observe distinct differences between datasets. 
While statistical baselines get average results on the TTRS and DHB datasets, they perform on par with other methods on TaFeng. The size of the datasets might be a factor contributing to these differences, as both TTRS and DHB have several times more interactions than TaFeng (\autoref{fig:data-comparison}).
In addition, absolute methods' performance on the TTRS dataset is generally higher than TaFeng and DHB.
The only exception is MF-based approaches (NMF, SVD, ALS), which could hardly achieve competitive results on this task over all datasets. 
On the other hand, in comparison with the alternatives, even RepeatNet results can be considered minor.
Finally, KNN-, I2I-,  VAE-based and sequential methods achieve average performance across all the datasets, with top-performing anomalies in different datasets.

\section{Discussion and Future Work}
\label{sec:discussion}

Extending recent advances in NPR applications \cite{Zhang2020HowTR, multitourrecommendations, multiperiodbasket}, this article aims to compare popular RecSys methods on the NPR task with publicly available datasets and provide an appropriate starting point for future research studies. Our exploratory data analysis showed a strong repetitive consumption trend over the reviewed datasets (\autoref{fig:data-scores} (b)), which, according to our benchmark results (\autoref{tab:benchmark-ml}), was often too difficult to generalize for evaluated RecSys approaches. On the other hand, statistical baselines, such as TopPersonal, and repeat-aware methods, such as RepeatNet, showed a substantial advantage over their alternatives in the benchmark. Finally, with such repetitive nature of the data, recommenders' performance on novel item prediction is still an open question, according to our preliminary comparison (\autoref{tab:benchmark-ml-new}).

Regarding repetitive consumption, our findings converge with previous studies \cite{nbr22, recanet22} on the importance of detailed RecSys methods evaluation on a subset of novel items, where one could see a distinct performance difference between overall (\autoref{tab:benchmark-ml}) and novel (\autoref{tab:benchmark-ml-new}) subsets. In addition, concerning recent works in frequency-aware RecSys approaches \cite{tifuknn, repeatnet, recanet22}, our results support the importance of model repeat-awareness when it comes to repetitive user consumption in order to achieve better performance. 
For example, missing in prior comparisons \cite{nbr22, recanet22}, RepeatNet shows promising results, according to our evaluations.
We speculate that the "repeat" mode of the proposed network helps generalize the datasets’ repetitive nature.
Finally, following recent advances in RecSys evaluation \cite{nbr22, arewereallymakingmuchprogress, popularbaselines}, our benchmark results (\autoref{tab:benchmark-ml}) revise that statistical baseline could be a competitive alternative to ML-based solutions\footnote{when it comes to strong repetitive patterns in data}. But most importantly, our findings offer a novel perspective on the NPR task, evaluating models that were never evaluated together before and using publicly-available datasets that were missed in previous NPR works \cite{Zhang2020HowTR, multitourrecommendations, multiperiodbasket}.

The main limitation of our results is the strong repetitive consumption trend found across revisited datasets\footnote{except TaFeng}. Although this dataset bias cannot be ruled out entirely, it is important to interpret our results together with the findings from prior research. Specifically, many recent studies on time-aware datasets, which we were focused on in this study, also found such biasing patterns \cite{nbr22, recanet22}. Taking this into account, the most plausible explanation of such bias is that these patterns are natural feature of this data rather than bias. 
% Taking this into account, the most plausible explanation is that it is not bias, but that such patterns are a natural feature of the actual data.

\section{Conclusion}

In this paper, we proposed a large-scale financial transactions dataset named TTRS that is based on user-merchant interactions and includes users' overall financial activity. Using the proposed dataset and closely related TaFeng and Dunnhumby ones, we evaluated various RecSys methods on next-period prediction tasks and compared their performance on common and novel item subsets. As shown by the benchmark, the user consumption repeatability factor is ubiquitous in many real-world datasets and challenging for many RecSys methods. At the same time, considering such repeatability, model efficiency in finding completely novel items for recommendation can still be considered questionable, providing avenues for future research.

\bibliography{main}

\clearpage
\appendix
\section{Appendix}

\begin{table*}[!h]
\footnotesize
    \begin{tabular}{|llll|}
        \hline
        \textbf{column}              & \textbf{description}                    & \textbf{\# unique values} & \textbf{column type} \\
        \hline
        party\_rk                    & unique user identification (anonymized) & 50000                     & int                  \\
        financial\_account\_type\_cd & account type                            & 2                         & categorical int      \\
        transaction\_type\_desc      & transaction type                        & 4                         & categorical string   \\
        merchant\_type               & merchant type (anonymized)              & 464                       & categorical int      \\
        merchant\_group\_rk          & merchant group identifier (anonymized)  & 2873                      & categorical int      \\
        category                     & merchant category                       & 36                        & categorical string  \\
        transaction\_dttm            & transaction timestamp                    & -                         & datetime             \\
        transaction\_amt        & transaction amount            & -                         & float                \\
        \hline
    \end{tabular}
    \caption{TTRS dataset description. For our experiments, we use $party\_rk$ as a user identifier and $merchant\_group\_rk$ as an item identifier.}
    \label{tab:DATASETDESCRIPTION}
\end{table*}

\begin{table*}[!h]
	\centering
	\caption{Hyperparameters search space for statistical and general top-n methods.}
	\vskip 0.15in
	\scalebox{0.99}{\begin{tabular}{cc}
		\toprule
		\textbf{Algorithms} & \textbf{Search Space}
		\\
		\midrule
		\textbf{Top Popular} & \begin{tabular}[c]{@{}c@{}}
		popular preprocessing $\in$ \{none, binary, log\}\\
	        \end{tabular}
	    \\
		\midrule
		\textbf{TopPersonal} & 
		\begin{tabular}[c]{@{}c@{}}
		minimal frequency $\in [0, 20]$\\
		personal preprocessing $\in$ \{none, binary, log\}\\
		\end{tabular}
		\\
		\midrule
		\textbf{NMF} & 
		\begin{tabular}[c]{@{}c@{}}
		solver $\in$ \{coordinate descent, multiplicative update\}\\
		initialization type $\in$ \{random, nndsvda\}\\
		$\beta \in$ \{frobenius, kullback-leibler\}\\
		number of factors $\in [1, 350]$\\
        l1 ratio $\in [0.0, 1.0]$\\
		\end{tabular}
		\\
		\midrule
		\textbf{SVD} & 
		\begin{tabular}[c]{@{}c@{}}
		hidden dimension $\in [32, 600]$\\
		latent dimension $\in [32, 600]$\\
		$\gamma \in [0.005, 0.0035, 0.01]$\\
		$\beta \in [0.2, 1.0]$\\
		dropout probability $\in [0.05, 0.5]$\\
		\end{tabular}
		\\
		\midrule
		\textbf{ALS} & 
		\begin{tabular}[c]{@{}c@{}}
		confidence scaling $\in$ \{True, False\}\\
		number of factors $\in [1, 200]$\\
		$\alpha \in [1e-3, 50.0]$\\
		$\epsilon \in [1e-3, 10.0]$\\
		regularization $\in [1e-5, 1e-2]$\\
		\end{tabular}
		\\
		\midrule
		\textbf{EASE} & 
		\begin{tabular}[c]{@{}c@{}}
		normalize matrix $\in$ \{True, False\}\\
		l2 norm $\in [0.0, 1.0]$\\
		\end{tabular}
		\\
		\midrule
		\textbf{SLIM} & 
		\begin{tabular}[c]{@{}c@{}}
		l1 ratio $\in [1e-5, 1.0]$\\
		$\alpha \in [1e-3, 1.0]$\\
		positive only $\in$ \{True, False\}\\
		top k $\in [5, 800]$\\
		\end{tabular}
	    \\
		\midrule
		\textbf{MultiVAE} & 
		\begin{tabular}[c]{@{}c@{}}
		total anneal steps $\in [10000, 500000]$\\
		mlp hidden size $\in [32, 600]$\\
		latent dimension $\in [32, 600]$\\
		encoder layers $\in [1, 4]$\\
		dropout prob $\in [0.05, 0.5]$\\
		$\beta \in [0.2, 1]$\\
		\end{tabular}
		\\
		\midrule
		\textbf{MacridVAE} & 
        \begin{tabular}[c]{@{}c@{}}
	    total anneal steps $\in [10000, 500000]$\\
		encoder hidden size$\in [32, 600]$\\
		embedding size$\in [32, 600]$\\
		encoder layers $\in [1, 4]$\\
		anneal cap $\in [0.2, 1]$\\
		kfac $\in [1, 20]$\\
		dropout prob $\in [0.05, 1]$\\
		reg weights $\in [1e-12, 1]$\\
		std $\in [0.055, 0.5]$\\
		nogb  $\{true, false\}$\\
		$\gamma \in \{0.0035,0.005,0.01\}$\\
		\end{tabular}
		\\
		\midrule
		\textbf{RecVAE} & 
		\begin{tabular}[c]{@{}c@{}}
		hidden dimension $\in [32, 600]$\\
		latent dimension$\in [32, 600]$\\
		dropout prob $\in [0.05, 0.5]$\\
		$\beta \in [0.2, 1]$\\
		$\gamma \in \{0.0035,0.005,0.01\}$\\
		\end{tabular}
		\\
		\bottomrule
	\end{tabular}}
	\label{tab:hyperparams}
\end{table*}

\begin{table*}[!h]
	\centering
	\caption{Hyperparameters search space for sequential and KKN-based methods.}
	\vskip 0.15in
	\scalebox{0.99}{\begin{tabular}{cc}
		\toprule
		\textbf{Algorithms} & \textbf{Search Space}
		\\	
		\midrule
		\textbf{GRU4Rec} & 
		\begin{tabular}[c]{@{}c@{}}
		embedding size$\in [32, 128]$\\
		hidden size $\in [32, 128]$\\
		num layers $\in [1, 4]$\\
		dropout prob $\in [0.05, 0.5]$\\
		\end{tabular}	
	    \\	
		\midrule
		\textbf{SASRec} & 
		\begin{tabular}[c]{@{}c@{}}
		num layers $\in [1, 4]$\\
		num heads $\in [1, 4]$\\
		hidden size $\in [8, 32]$\\
		inner size $\in [32, 128]$\\
		hidden act  $\{gelu, relu, swish, tanh, sigmoid\}$\\
		hidden dropout prob $\in [0.05, 0.5]$\\
		attn dropout prob $\in [0.05, 0.5]$\\
		layer norm eps $\in [1e-12, 1e-6]$\\
		initializer range $\in [0.0, 0.02]$\\
		\end{tabular}	
		\\	
		\midrule
		\textbf{BERT4Rec} & 
		\begin{tabular}[c]{@{}c@{}}
		num layers $\in [1, 4]$\\
		num heads $\in [1, 4]$\\
		hidden size $\in [8, 32]$\\
		inner size $\in [32, 128]$\\
		hidden act  $\{gelu, relu, swish, tanh, sigmoid\}$\\
		hidden dropout prob $\in [0.05, 0.5]$\\
		attn dropout prob $\in [0.05, 0.5]$\\
		layer norm eps $\in [1e-12, 1e-6]$\\
		mask ration $\in [0.05, 0.8]$\\
		initializer range $\in [0.0, 0.02]$\\
		\end{tabular}
		\\	
		\midrule
		\textbf{RepeatNet} & 
		\begin{tabular}[c]{@{}c@{}}
		embedding size$\in [128]$\\
		hidden size $\in [8, 32]$\\
		inner size $\in [32, 128]$\\
		hidden act  $\{gelu, relu, swish, tanh, sigmoid\}$\\
		hidden dropout prob $\in [0.05, 0.5]$\\
		attn dropout prob $\in [0.05, 0.5]$\\
		layer norm eps $\in [1e-12, 1e-6]$\\
		mask ration $\in [0.05, 0.8]$\\
		initializer range $\in [0.0, 0.02]$\\
		\end{tabular}
		\\
		\midrule
		\textbf{ItemKNN} & 
		\begin{tabular}[c]{@{}c@{}}
		top k $\in [1, 200]$\\
		shrink $\in [0, 600]$\\
		similarity $\in$ \{cosine, jaccard, asymmetric, dice, tversky\}\\
		\end{tabular}
		\\
		\midrule
		\textbf{TifuKNN} & 
		\begin{tabular}[c]{@{}c@{}}
		rb $\in [0.1, 0.9]$\\
		rg $\in [0.1, 0.9]$\\
		m $\in [2, 23]$\\
		alpha $\in [0.0, 1.0]$\\
		nn number $\in [100, 1300]$\\
		\end{tabular}
		\\
		\bottomrule
	\end{tabular}}
	\label{tab:hyperparams}
\end{table*}

\begin{table}
\caption{Next-month recommendation benchmark. Ground truth test interactions are not necessarily new to users, and recommendation lists are unfiltered. All metrics are averaged over 2 (TaFeng) or 6 (TTRS and DHB) test folds and normalized to the $[0, 100]$ values range for presentation convenience.
The top three models are marked in \textbf{bold}.
% The best model is marked in \topf{bold}, the second best model is \tops{underlined} and the third best model is in \topt{italics}.
}
% \label{tab:benchmark-ml}
\begin{center}
% \begin{small}
\begin{sc}
\footnotesize
\begin{tabular}{ll|ccc|ccc|ccc}
\toprule

\multicolumn{2}{l|}{Dataset} & \multicolumn{3}{c|}{TaFeng} & \multicolumn{3}{c|}{TTRS} & \multicolumn{3}{c}{DHB} \\
% \cmidrule(r){3-5}
% \cmidrule(r){6-8}
% \cmidrule{9-11}
$@K$ & Method & Recall & MAP & NDCG & Recall & MAP & NDCG & Recall & MAP & NDCG \\
\midrule
\multirow{16}{*}{20} & toppopular & 8.33  & 2.93  & 7.77  & 45.43  & 21.79  & 38.93  & 13.28  & 10.66  & 24.20 \\
& toppersonal & 16.05 & 7.90 & 16.57 & 69.82 & 52.08 & \topt{68.22} & \tops{29.79} & \tops{30.94} & \tops{48.60}\\
\cmidrule{2-11}
& nmf & 14.88  & 6.90 & 15.04 & 64.52  & 49.67 & 65.28 & 25.96 & 26.65 & 44.36\\
& svd & 17.55 & 7.78 & 16.92 & 64.41  & 49.83 & 65.35 & 27.47 & 28.03 & 45.96\\
& ALS & 18.57 & 6.80 & 15.96 & 66.92 & 42.59  & 59.47  & 25.09 & 21.32 & 38.60\\
\cmidrule{2-11}
& ease & \topt{18.85} & \tops{8.28} & \topt{17.75} & 55.60  & 29.23  & 47.49  & 22.82 & 17.88  & 35.18 \\
& slim & 16.71 & 7.44 & 16.05 & 51.59  & 26.40  & 44.21  & 18.80  & 14.44  & 30.38 \\
\cmidrule{2-11}
& multivae & 15.79 & 5.59  & 13.55  & 68.06 & 45.82  & 62.78  & 23.00 & 19.24 & 36.40\\
& macridvae & 11.58  & 4.21  & 10.51  & 51.71  & 29.52  & 46.97  & 13.87  & 10.94  & 24.67 \\
& recvae & 18.51 & \topt{8.26} & 17.63 & \topf{70.53} & \tops{52.66} & \tops{68.63} & \topt{29.41} & \topt{29.49} & \topt{47.35}\\
\cmidrule{2-11}
& gru4rec & 7.25  & 2.10  & 6.00  & 65.37 & 49.04 & 65.20 & 19.62  & 16.87  & 33.32 \\
& sasrec & 11.14  & 4.62  & 10.55  & 67.54 & 50.21 & 66.30 & 21.00  & 17.99  & 34.67 \\
& bert4rec & 10.81  & 4.32  & 10.16  & 66.52 & 49.26 & 65.45 & 21.89  & 19.68 & 36.89\\
% \cmidrule{2-11}
& repeatnet & \topf{19.00} & \topf{8.33} & \topf{17.98} & \tops{70.14} & \topf{53.16} & \topf{69.01} & \topf{31.41} & \topf{32.50} & \topf{50.49}\\
\cmidrule{2-11}
& itemknn & 13.88  & 5.24  & 12.31  & 45.04  & 20.51  & 37.28  & 15.42  & 11.44  & 25.79 \\
& tifuknn & \tops{18.89} & \topt{8.26} & \tops{17.79} & \topt{69.94} & \topt{52.09} & 68.21 & 28.74 & 27.98 & 46.06\\
\midrule 

\multirow{16}{*}{50} & toppopular & 13.68  & 3.16  & 9.75  & 60.50  & 24.10  & 44.93  & 20.04  & 8.26  & 22.67 \\
& toppersonal & 22.34 & 8.20 & 18.79 & 78.83 & 54.17 & 72.07 & \topt{41.13} & \tops{25.17} & \topt{45.34}\\
\cmidrule{2-11}
& nmf & 18.69  & 7.03 & 16.29 & 67.39  & 50.17 & 66.41 & 32.30 & 20.06 & 38.64\\
& svd & 21.67 & 7.95 & 18.29 & 66.64  & 50.21 & 66.23  & 34.56 & 21.48 & 40.53\\
& ALS & 23.75 & 7.14 & 17.84 & 71.73  & 43.56  & 61.45  & 35.40 & 17.27 & 36.36\\
\cmidrule{2-11}
& ease & 25.61 &\topt{ 8.73} & 20.24 & 73.56 & 32.48  & 54.61  & 36.53 & 15.86  & 35.79\\
& slim & 24.82 & 7.98 & 19.11 & 69.54  & 29.39  & 51.29  & 29.15  & 12.22  & 30.04 \\
\cmidrule{2-11}
& multivae & 22.56 & 6.01  & 16.05  & \tops{80.18} & 48.50  & 67.86 & 34.23 & 16.19 & 35.42\\
& macridvae & 18.58  & 4.58  & 13.12  & 68.83  & 32.47  & 53.85  & 20.94  & 8.52  & 23.22 \\
& recvae & \topt{25.65} &\tops{ 8.78} & \topt{20.33} & \topf{80.91} & \tops{55.05} & \topf{73.02} & 40.95 & 24.08 & 44.42\\
\cmidrule{2-11}
& gru4rec & 12.88  & 2.37  & 8.14  & 77.28 & 51.51 & 70.18 & 28.00  & 14.03  & 31.71 \\
& sasrec & 17.68  & 4.97  & 12.99  & 79.33 & 52.77 & 71.25 & 29.78  & 15.06  & 33.09 \\
& bert4rec & 17.29  & 4.65  & 12.57  & 78.39 & 51.78 & 70.41 & 31.00  & 16.22 & 34.88 \\
% \cmidrule{2-11}
& repeatnet & \tops{25.80} &\topt{ 8.73} & \tops{20.42} & 79.03 & \topf{55.21} & \tops{72.79} & \topf{44.55} & \topf{27.20} & \topf{48.05}\\
\cmidrule{2-11}
& itemknn & 20.36  & 5.63  & 14.70  & 65.87  & 23.70  & 45.32  & 24.63  & 9.37  & 25.37 \\
& tifuknn & \topf{26.84} &\topf{ 8.85} & \topf{20.83} & \topt{79.54} & \topt{54.31} & \topt{72.30} & \tops{43.84} & \topt{24.58} & \tops{45.81}\\

\midrule

\multirow{16}{*}{100} &toppopular & 20.98  & 3.44  & 12.28  & 73.03  & 25.14  & 48.97  & 26.79  & 8.80  & 25.73 \\
& toppersonal & 28.94 & 8.52 & 21.09 & 85.23 & 54.89 & 74.28 & 47.97 & \tops{26.14} & \topt{48.43}\\
\cmidrule{2-11}
& nmf & 22.11  & 7.17 & 17.47  & 69.80  & 50.38 & 67.24 & 37.23  & 20.46 & 40.68\\
& svd & 24.89  & 8.10 & 19.42 & 68.72  & 50.40 & 66.95  & 39.32 & 21.88 & 42.48\\
& ALS & 27.44 & 7.32 & 19.12 & 75.60  & 43.95  & 62.79  & 42.96 & 18.18 & 39.72\\
\cmidrule{2-11}
& ease & 29.90 & 8.96 & 21.75 & 81.98 & 33.33  & 57.37  & 46.45 & 17.30 & 40.38\\
& slim & 30.02 & 8.25 & 20.94 & 80.88  & 30.49  & 55.03  & 39.03  & 13.36  & 34.58 \\
\cmidrule{2-11}
& multivae & 28.20 & 6.29  & 18.02 & \tops{87.25} & 49.33  & 70.31 & 43.53 & 17.40 & 39.67\\
& macridvae & 26.31 & 4.93  & 15.82  & 79.46  & 33.50  & 57.39  & 27.80  & 9.09  & 26.32 \\
& recvae & \tops{30.96} &\tops{ 9.04} & \topt{22.14} & \topf{87.40} & \tops{55.82} & \topf{75.28} & \tops{49.28} & 25.28 & 48.15\\
\cmidrule{2-11}
& gru4rec & 19.21  & 2.59  & 10.30  & 85.44 & 52.42 & 73.01 & 36.11  & 14.87  & 35.41 \\
& sasrec & 24.95  & 5.26  & 15.45  & \topt{86.93} & 53.65 & 73.90 & 38.00  & 15.97  & 36.86 \\
& bert4rec & 24.29  & 4.94  & 14.98  & 86.19 & 52.67 & 73.13 & 39.46 & 17.19  & 38.74 \\
% \cmidrule{2-11}
& repeatnet & \topt{30.90} &\topt{ 8.98} & \tops{22.18} & 84.74 & \topf{55.86} & \tops{74.78} & \topt{48.54} & \topf{27.62} & \topf{49.63}\\
\cmidrule{2-11}
& itemknn & 25.72  & 5.87  & 16.55  & 79.62  & 24.94  & 49.74  & 33.66  & 10.26  & 29.47 \\
& tifuknn & \topf{33.47} &\topf{ 9.21} & \topf{23.13} & 85.56 & \topt{55.00} & \topt{74.38} & \topf{51.88} & \topt{25.85} & \tops{49.49}\\

\bottomrule
\end{tabular}
\end{sc}
% \end{small}
\end{center}
\end{table}

\begin{table}
\caption{Next-month novel items recommendation benchmark. Ground truth objects are new to users, and recommendation lists do not contain items interacted with during training. All metrics are averaged over 2 (TaFeng) or 6 (TTRS and DHB) test folds and normalized to the $[0, 100]$ values range for presentation convenience. 
The top three models are marked in \textbf{bold}.
% The best model is marked in \topf{bold}, the second best model is \tops{underlined} and the third best model is in \topt{italics}.
}
% \label{tab:benchmark-ml}
\begin{center}
% \begin{small}
\begin{sc}
\footnotesize
\begin{tabular}{ll|ccc|ccc|ccc}
\toprule

\multicolumn{2}{l|}{Dataset} & \multicolumn{3}{c|}{TaFeng} & \multicolumn{3}{c|}{TTRS} & \multicolumn{3}{c}{DHB} \\
% \cmidrule(r){3-5}
% \cmidrule(r){6-8}
% \cmidrule{9-11}
$@K$ & Method & Recall & MAP & NDCG & Recall & MAP & NDCG & Recall & MAP & NDCG \\
\cmidrule{1-11}
\multirow{16}{*}{20} & toppopular &\topt{ 6.83 } &\tops{ 2.59 } &\tops{ 6.34 } & 30.02  & 10.19  & 19.78  & 7.37  & 2.93  & 8.74 \\
& toppersonal & 6.77  &\tops{ 2.59 } &\topt{ 6.32 } & 30.03  & 10.25  & 19.84  & 7.34  & 2.90  & 8.68 \\
\cmidrule{2-11}
& nmf & 2.17  & 0.65  & 1.90  & 5.24  & 1.47  & 3.29  & 2.80  & 0.81  & 2.95 \\
& svd & 2.35  & 0.64  & 1.94  & 3.73  & 1.03  & 2.38  & 2.64  & 0.78  & 2.82 \\
& als & 3.72  & 1.00  & 2.94  & 13.52  & 4.07  & 8.57  & 5.46  & 1.75  & 5.81 \\
\cmidrule{2-11}
& ease & 5.97  & 1.91  & 5.03  & 33.25  & \topt{12.31 } & 22.59  & 6.92  & 2.55  & 7.87 \\
& slim & 5.73  & 1.64  & 4.63  & 32.99  & 11.80  & 22.04  &\tops{ 8.10 } &\tops{ 3.03 } &\tops{ 9.11 }\\
\cmidrule{2-11}
& multivae & 4.52  & 1.13  & 3.44  & \topf{35.15 } & \topf{12.53 } & \topf{23.31 } & 7.53  & 2.75  & 8.45 \\
& macridvae &\topf{ 7.50 } & 2.11  & 5.85  & 33.24  & 11.49  & 21.90  &\topt{ 7.67 } &\topt{ 3.00 } &\topt{ 9.00 }\\
& recvae & 5.82  & 1.74  & 4.81  & \topt{34.79 } & \tops{12.50 } & \tops{23.21 } & 7.20  & 2.77  & 8.36 \\
\cmidrule{2-11}
& gru4rec & 4.89  & 1.07  & 3.50  & 32.06  & 10.48  & 20.40  & 6.38  & 2.13  & 6.88 \\
& sasrec & 5.02  & 1.28  & 3.78  & \tops{35.02 } & 12.06  & \topt{22.72 } & 6.46  & 2.13  & 6.88 \\
& bert4rec & 5.11  & 1.24  & 3.77  & 33.56  & 11.30  & 21.57  & 7.36  & 2.59  & 8.09 \\
% \cmidrule{2-11}
& repeatnet & 6.07  & 1.33  & 4.28  & 31.09  & 10.51  & 20.36  & 3.49  & 1.20  & 4.12 \\
\cmidrule{2-11}
& itemknn & 4.39  & 1.23  & 3.48  & 33.96  & 12.00  & 22.58  &\topf{ 8.28 } &\topf{ 3.23 } &\topf{ 9.52 }\\
& tifuknn &\tops{ 6.97 } &\topf{ 2.62 } &\topf{ 6.43 } & 31.62  & 10.87  & 20.80  & 7.46  & 2.94  & 8.77 \\
\midrule 

\multirow{16}{*}{50} & toppopular & 12.06  &\tops{ 2.83 } &\topt{ 8.28 } & 45.43  & 11.25  & 24.42  & 12.57  & 2.87  & 10.28 \\
& toppersonal & \topt{12.14 } &\tops{ 2.83 } &\tops{ 8.31 } & 45.41  & 11.30  & 24.47  & 12.49  & 2.84  & 10.20 \\
\cmidrule{2-11}
& nmf & 3.86  & 0.71  & 2.54  & 9.54  & 1.64  & 4.55  & 5.57  & 0.84  & 3.96 \\
& svd & 4.67  & 0.72  & 2.80  & 7.08  & 1.15  & 3.39  & 5.29  & 0.81  & 3.80 \\
& ALS & 7.07  & 1.13  & 4.16  & 22.79  & 4.52  & 11.33  & 10.28  & 1.83  & 7.52 \\
\cmidrule{2-11}
& ease & 9.93  & 2.07  & 6.48  & 48.08  & 13.36  & 27.05  & 11.76  & 2.54  & 9.36 \\
& slim & 10.14  & 1.83  & 6.27  & 49.69  & 12.98  & 27.07  & \tops{14.11 } &\tops{ 3.07 } & \tops{11.03 }\\
\cmidrule{2-11}
& multivae & 8.65  & 1.30  & 4.97  & \tops{53.58 } & \topf{13.88 } & \topf{28.86 } & \topt{13.37 } & 2.80  & 10.35 \\
& macridvae & \topf{13.24 } & 2.39  & 7.99  & 51.89  & 12.86  & 27.53  & 13.09  &\topt{ 2.96 } & \topt{10.60 }\\
& recvae & 10.23  & 1.93  & 6.45  & \topt{53.13 } & \tops{13.84 } & \tops{28.72 } & 12.50  & 2.77  & 10.02 \\
\cmidrule{2-11}
& gru4rec & 9.47  & 1.26  & 5.19  & 50.77  & 11.79  & 26.02  & 11.78  & 2.20  & 8.72 \\
& sasrec & 9.83  & 1.49  & 5.54  & \topf{53.75 } & \topt{13.42 } & \topt{28.35 } & 12.08  & 2.23  & 8.83 \\
& bert4rec & 9.97  & 1.44  & 5.56  & 52.08  & 12.62  & 27.13  & 13.33  & 2.67  & 10.09 \\
% \cmidrule{2-11}
& repeatnet & 10.27  & 1.51  & 5.84  & 45.93  & 11.53  & 24.85  & 6.20  & 1.15  & 4.94 \\
\cmidrule{2-11}
& itemknn & 8.03  & 1.38  & 4.81  & 52.55  & 13.37  & 28.19  & \topf{14.22 } &\topf{ 3.22 } & \topf{11.32 }\\
& tifuknn & \tops{12.44 } &\topf{ 2.86 } &\topf{ 8.44 } & 47.94  & 12.00  & 25.70  & 12.86  & 2.91  & 10.41 \\

\midrule

\multirow{16}{*}{100} & toppopular & 18.80  &\topt{ 3.04 } & \topt{10.42 } & 61.47  & 11.85  & 28.37  & 18.59  & 3.16  & 12.68 \\
& toppersonal & \topt{18.97 } &\tops{ 3.05 } & \tops{10.48 } & 61.16  & 11.89  & 28.34  & 18.56  & 3.13  & 12.61 \\
\cmidrule{2-11}
& nmf & 6.38  & 0.76  & 3.35  & 15.11  & 1.75  & 5.90  & 9.36  & 0.95  & 5.46 \\
& svd & 7.80  & 0.79  & 3.81  & 11.49  & 1.23  & 4.49  & 9.08  & 0.92  & 5.29 \\
& ALS & 11.14  & 1.22  & 5.45  & 32.46  & 4.77  & 13.71  & 16.21  & 2.07  & 9.85 \\
\cmidrule{2-11}
& ease & 14.41  & 2.19  & 7.92  & 59.23  & 13.79  & 29.83  & 17.06  & 2.77  & 11.47 \\
& slim & 15.05  & 1.96  & 7.83  & 63.53  & 13.53  & 30.50  & \topf{20.98 } &\tops{ 3.41 } & \tops{13.74 }\\
\cmidrule{2-11}
& multivae & 13.73  & 1.44  & 6.61  & \tops{68.66 } & \topf{14.51 } & \topf{32.59 } & 20.26  & 3.13  & 13.07 \\
& macridvae & \topf{20.36 } & 2.63  & 10.30  & 66.81  & 13.47  & 31.23  & 19.30  &\topt{ 3.26 } & \topt{13.09 }\\
& recvae & 16.03  & 2.09  & 8.26  & \topt{67.95 } & \tops{14.46 } & \tops{32.41 } & 18.73  & 3.05  & 12.49 \\
\cmidrule{2-11}
& gru4rec & 15.05  & 1.41  & 6.96  & 66.45  & 12.42  & 29.91  & 18.40  & 2.49  & 11.32 \\
& sasrec & 16.21  & 1.66  & 7.52  & \topf{68.81 } & \topt{14.05 } & \topt{32.11 } & 18.81  & 2.53  & 11.46 \\
& bert4rec & 15.91  & 1.61  & 7.47  & 67.39  & 13.24  & 30.93  & \topt{20.46 } & 3.01  & 12.89 \\
% \cmidrule{2-11}
& repeatnet & 15.64  & 1.66  & 7.56  & 59.51  & 12.05  & 28.21  & 10.31  & 1.28  & 6.56 \\
\cmidrule{2-11}
& itemknn & 12.57  & 1.49  & 6.24  & 67.91  & 14.01  & 31.99  & \tops{20.75 } &\topf{ 3.54 } & \topf{13.89 }\\
& tifuknn & \tops{19.87 } &\topf{ 3.10 } & \topf{10.81 } & 62.26  & 12.55  & 29.24  & 18.72  & 3.18  & 12.73 \\

\bottomrule
\end{tabular}
\end{sc}
% \end{small}
\end{center}
\end{table}

% \begin{figure}[!htbp] %[t!]
% \centerline{\includesvg[width=0.5\textwidth]{figures/preprocessing_map}}
% \caption{Validation model performance ($MAP@10$) relative to the preprocessing used. TaFeng dataset is in the left column, TTRS in the middle, and DHB in the right.
% Each row corresponds to one of the evaluated methods, from top to bottom: NMF, SVD, EASE, SLIM, MultiVAE, MultiDAE, RecVAE.}
% \label{fig:map-preprocessing}
% \end{figure}

\end{document}